\RequirePackage{fix-cm}
\documentclass[twocolumn]{svjour3}          
\smartqed  
\usepackage{graphicx}
\usepackage{amsmath}
\usepackage{amssymb}
\usepackage{natbib}
\usepackage{pifont}

\usepackage{subfigure}
\usepackage{bm}
\usepackage{bbm}
\usepackage{algorithmic}
\usepackage{algorithm}
\usepackage{multirow}
\usepackage{array}
\usepackage{ marvosym }
\usepackage{authblk}

\newcolumntype{P}[1]{>{\centering\arraybackslash}p{#1}}
\newcolumntype{M}[1]{>{\centering\arraybackslash}m{#1}}
\newcommand{\ie}{\textit{i.e.}}
\newcommand{\wrt}{\textit{w.r.t.}}
\newcommand{\eg}{\textit{e.g.}}

\usepackage[pagebackref=true,breaklinks=true,letterpaper=true,colorlinks,bookmarks=false]{hyperref}

 \journalname{International Journal of Computer Vision}

\begin{document}

\title{A Shape Transformation-based Dataset Augmentation Framework for Pedestrian Detection
}


%
\authorrunning{Chen \textit{et al.}} 
\author{Zhe Chen $^1$ \and
 Wanli~Ouyang $^2$\and
Tongliang~Liu $^1$\and
Dacheng~Tao $^1$
}
\institute{\Letter~Zhe Chen  (zhe.chen1@sydney.edu.au)\\
	Wanli Ouyang (wanli.ouyang@sydney.edu.au)\\
	Tongliang Liu (tongliang.liu@sydney.edu.au)\\
	Dacheng Tao (dacheng.tao@sydney.edu.au)\\
	1 UBTECH Sydney Artificial Intelligence Centre, The University of Sydney, Sydney, Australia\\
	2 The University of Sydney, Sydney, Australia
}

\maketitle

\begin{abstract}
Deep learning-based computer vision is usually data hungry. Many researchers attempt to augment datasets with synthesized data to improve model robustness. However, the augmentation of popular pedestrian datasets, such as Caltech and Citypersons, can be extremely challenging because real pedestrians are commonly in low quality. Due to the factors like occlusions, blurs, and low-resolution, it is significantly difficult for existing augmentation approaches, which generally synthesize data using 3D engines or generative adversarial networks (GANs), to generate realistic-looking pedestrians. Alternatively, to access much more natural-looking pedestrians, we propose to augment pedestrian detection datasets by transforming real pedestrians from the same dataset into different shapes. Accordingly, we propose the Shape Transformation-based Dataset Augmentation (STDA) framework. The proposed framework is composed of two subsequent modules, \ie ~the shape-guided deformation and the environment adaptation. In the first module, we introduce a shape-guided warping field to help deform the shape of a real pedestrian into a different shape. Then, in the second stage, we propose an environment-aware blending map to better adapt the deformed pedestrians into surrounding environments, obtaining more realistic-looking pedestrians and more beneficial augmentation results for pedestrian detection. 
Extensive empirical studies on different pedestrian detection benchmarks show that the proposed STDA framework consistently produces much better augmentation results than other pedestrian synthesis approaches using low-quality pedestrians. By augmenting the original datasets, our proposed framework also improves the baseline pedestrian detector by up to 38\% on the evaluated benchmarks, achieving state-of-the-art performance.
\end{abstract}

\section{Introduction}

With the introduction of large-scale pedestrian datasets \citep{dollar2009pedestrian,dollar2012pedestrian,zhang2017citypersons,geiger2013vision}, deep convolutional neural networks (DCNNs) have achieved promising detection accuracy. However, the trained DCNN detectors may not be robust enough due to the issue that negative background examples greatly exceed positive foreground examples during training. Recent studies have confirmed that DCNN detectors trained with the limited foreground examples can be vulnerable to difficult objects which have unexpected states \citep{huang2017expecting} and diversified poses \citep{alcorn2018strike}. 

\begin{figure*}[t]
\begin{center}
\includegraphics[width=0.8\linewidth]{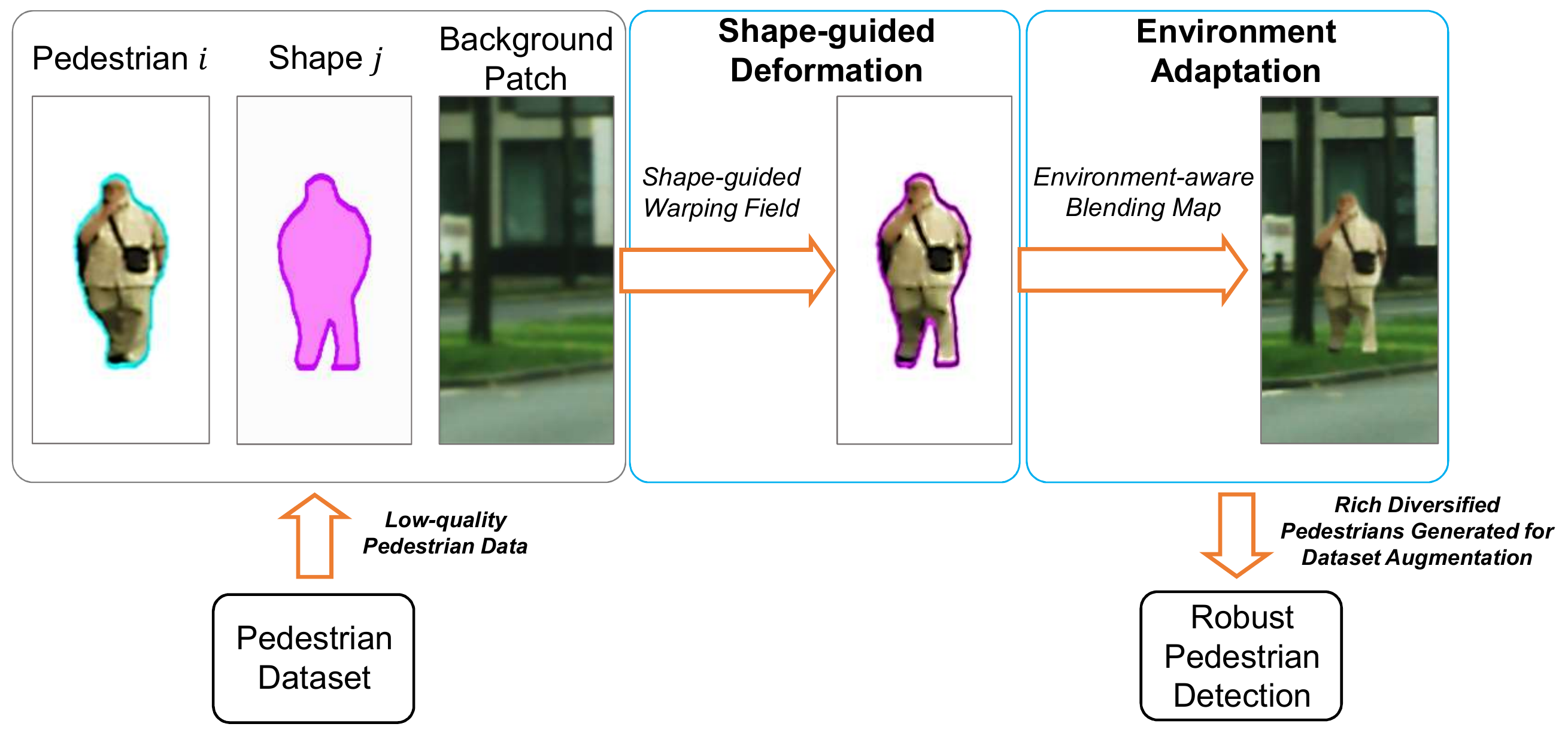}
\end{center}
   \caption{We propose the shape transformation-based dataset augmentation framework for pedestrian detection. In the framework, we subsequently introduce the shape-guided warping field to deform pedestrians and the environment-aware blending map to adapt the deformed pedestrians into background environments. Our proposed framework can effectively generate more realistic-looking pedestrians for augmenting pedestrian datasets in which real pedestrians are usually in low-quality. Best view in color.}
\label{fig:title}
\end{figure*}
To improve detector robustness, besides designing new machine learning algorithms, many researchers attempted to augment training datasets by generating new foreground examples. For instance, Huang \textit{et al.} \citep{huang2017expecting} used a 3D game engine to simulate pedestrians and adapt them into the pedestrian datasets. Other studies \citep{ma2018disentangled,siarohin2018deformable,zanfir2018human,ge2018fd} attempted to augment the person re-identification datasets by transferring the poses of pedestrians using generative adversarial networks (GANs). Despite progress, it is still very challenging to adequately apply existing augmentation approaches on the common pedestrian detection datasets. First, synthesizing pedestrians using external platforms like the 3D game engines may introduce a significant domain gap between synthesized pedestrians and real pedestrians, limiting the overall benefits for the generated pedestrians to improve the model robustness for detecting real pedestrians. Moreover, regarding the methods that utilize GANs to render pedestrians, they generally require rich appearance details from paired training images to help define the desired output of  generative networks during training procedures. However, in common pedestrian detection datasets like Caltech \citep{dollar2012pedestrian} and CityPersons \citep{zhang2017citypersons}, pedestrians are usually in low quality due to the factors like heavy occlusions, blurry appearance, and low-resolution caused by small sizes. As a result, these available real pedestrians only provide extremely limited amount of appearance details that can be used for training generative networks. Without sufficient description of the desired appearance of synthesized pedestrians, we can show in our experiments that current GAN-based methods only generate less realistic or even corrupted pedestrians using very low-quality pedestrians from common pedestrian detection datasets.

By addressing above issues, we propose to augment pedestrian datasets by transforming real pedestrians from the same dataset according to different shapes (\ie ~ segmentation masks in this study) rather than rendering new pedestrians. Our motivation comes from the following observations. First, unlike existing methods that require sufficient appearance details to define the desired output, it is much easier to access rich pixel-level shape deformation supervision which defines the deformation from a \textit{shape} to another shape, if only low-quality pedestrian examples are available in the datasets. The learned deformation between different shapes can guide the deformation of appearances of the real pedestrians, avoiding the requirement of detailed supervision information to directly define the transformed appearances. In addition, since the shape information can naturally distinguish foreground areas from background areas, we can simply focus on adapting synthesized foreground appearances into background environments, avoiding the rick of further generating unnatural background environments together with the synthesized pedestrians as required in current GAN-based approaches. Last but not the least, we find that transforming real pedestrians based on different shapes can effectively increase foreground sample diversity while still maintaining the appearance characteristics of real pedestrians adequately.

Based on these observations, we devise a Shape Transformation based Dataset Augmentation (STDA) framework to fulfill the pedestrian dataset augmentation task more effectively. In particular, the framework first deforms a real pedestrian into a similar pedestrian but with a different shape and then adapts the shape-deformed pedestrians into surrounding environments on the image to be augmented. In the STDA framework, we introduce a shape-guided warping field, which is a set of vectors that define the warping operation between shapes, to further define an appropriate deformation between the shapes and the appearances of the real pedestrians. 
Moreover, we introduce an environment-aware blending map to help the shape-deformed pedestrians better blend into various background environments, delivering more realistic-looking pedestrians on the image.

In this study, our key contributions are listed as follows:
\begin{itemize}
\item 
We propose a shape transformation-based dataset augmentation framework to augment the pedestrian detection datasets and improve pedestrian detection accuracy. 
To the best of our knowledge, we are the first that apply the shape-transformation-based data synthesis methodology for pedestrian detection.
\item 
We propose the shape-guided warping field to help define a proper shape deformation procedure. We also introduce an environment-aware blending map to better adapt the shape-transformed pedestrians into different backgrounds, achieving better augmentation results on the image.
\item 
We introduce a shape constraining operation to improve shape deformation quality and a novel hard positive mining loss to further magnify the benefits of the synthesized pedestrians for improving detection robustness. 
\item 
Our proposed framework is promising for generating pedestrians using low-quality examples.
Comprehensive evaluations on the famous Caltech \citep{dollar2012pedestrian} and CityPersons \citep{zhang2017citypersons} benchmarks validate that our proposed framework can generate more realistic-looking pedestrians than existing methods using low-quality data. With pedestrian datasets augmented by our framework, we promisingly boost the performance of the baseline pedestrian detector, accessing superior performance to other cutting-edge pedestrian detectors. 
\end{itemize}

\section{Related Work}

\subsection{Pedestrian Detection}
Pedestrian is critical in many applications such as robotics and autonomous driving \citep{enzweiler2008monocular,dollar2009pedestrian,dollar2012pedestrian,zhang2016far}. Traditional pedestrian detectors generally use hand-crafted features \citep{viola2005detecting,ran2007pedestrian} and adopt human part-based detection strategy \citep{felzenszwalb2010object}  or cascaded structures \citep{felzenszwalb2010cascade,bar2010part,felzenszwalb2008discriminatively}. 
Recently, by taking advantages of large-scale pedestrian datasets   \citep{dollar2009pedestrian,dollar2012pedestrian,zhang2017citypersons,geiger2013vision}, researchers have greatly improved the pedestrian detection performance with DCNNs \citep{simonyan2014very,he2016deep}. Among the DCNN detectors, two-stage detection pipelines \citep{ouyang2013joint,ren2015faster,li2018scale,cai2016unified, zhang2016faster,du2017fused} usually perform better than single-stage detection pipelines \citep{liu2016ssd,redmon2016you,lin2018graininess}. 
Despite progress, the issue that foreground and background examples are extremely unbalanced in pedestrian datasets still affects the robustness of the DCNN detectors adversely. Current pedestrian detectors could still be fragile to even small transformation of pedestrians. To tackle this problem, many researchers tend to augment the datasets by synthesizing new foreground data.

\subsection{Simulation-based Dataset Augmentation}
To achieve dataset augmentation, researchers have used 3D simulation platforms to synthesize new examples for the datasets. 
For example, 
\citep{lerer2016learning,ros2016synthia} used a 3D game engine to help build new datasets. More related studies used the 3D simulation platforms to augment pedestrian-related datasets. In particular, \citep{pishchulin2011learning,hattori2015learning} employed a game engine to synthesize training data for pedestrian detection. In addition, \citep{huang2017expecting} applied a GAN to narrow the domain gap between the 3D simulated pedestrians and the natural pedestrians to augment pedestrian datasets, but this method brings limited improvement on common pedestrian detection, suggesting that the domain gap is still large.
However, there is still a significant domain gap between simulated pedestrians and real pedestrians. Such gap could further pose negative effects on DCNN detectors, making the augmented datasets deliver incremental improvements on pedestrian detection. 

\subsection{GAN-based Dataset Augmentation}
Recently, with several improvements \citep{radford2015unsupervised,arjovsky2017wasserstein,gulrajani2017improved}, GANs\citep{goodfellow2014generative} have shown great benefits on synthesis-based applications such as image-to-image translation \citep{isola2017image,liu2017unsupervised,isola2017image,CycleGAN2017} and skeleton-to-image generation \citep{villegas2017learning,yan2017skeleton}. 

\begin{figure}[h]
\begin{center}
\includegraphics[width=\linewidth,height=0.17\textheight]{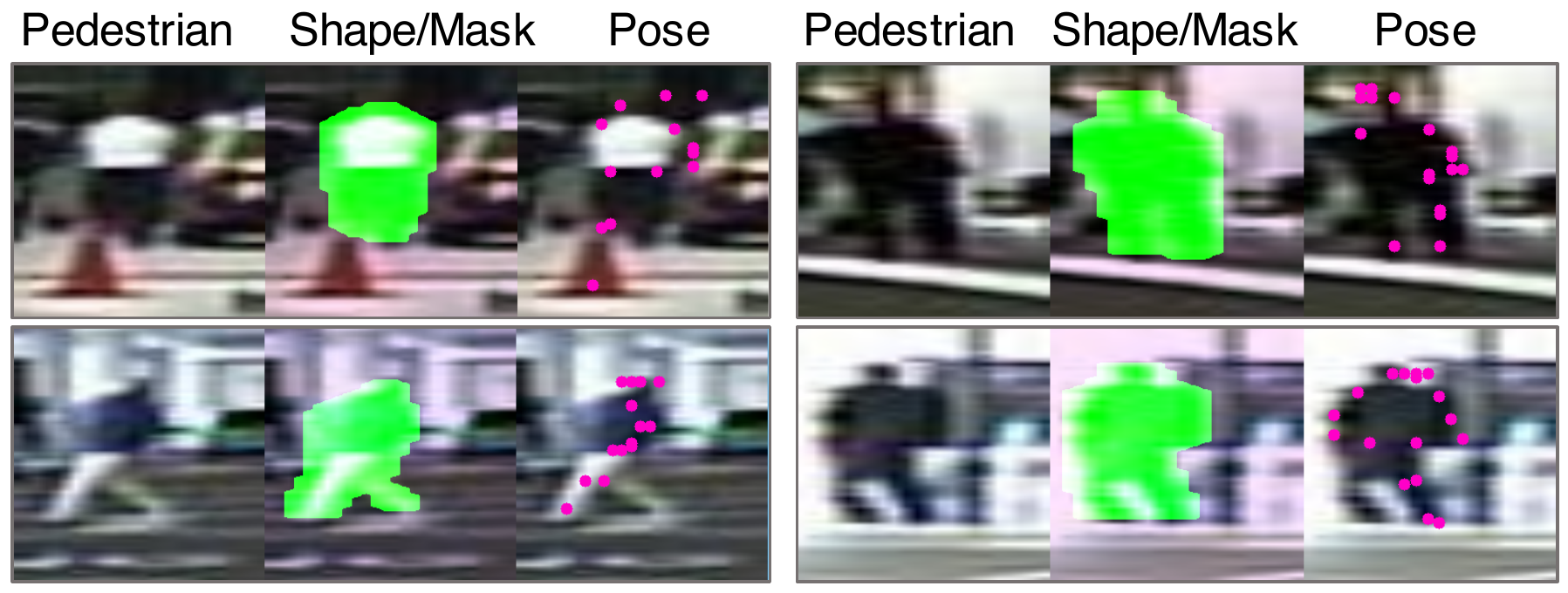}
\end{center}
   \caption{Some examples showing that the shape estimation results are more accurate than pose estimation results on low-quality images using the same Mask RCNN model. Best view in color.}
\label{fig:mask}
\end{figure}

In the literature of person re-identification task, many works attempted to transfer the poses of real pedestrians to deliver diversified pedestrians for the augmentation.
For instance, \citep{liu2018pose,ma2018disentangled,siarohin2018deformable, zanfir2018human,ge2018fd,zheng2017unlabeled,ma2017pose} introduced various techniques to transform the human appearance according to 2D or 3D poses and improve the person re-identification performance. In practice, these methods require accurate pose information or paired training images that contain rich appearance details to achieve successful transformation. However, existing widely used pedestrian datasets like Caltech provide neither pose annotations nor paired appearance information for training GANs. Furthermore, in current pedestrian datasets, a large number of small pedestrians whose appearances are usually in low quality can make existing pose estimators difficult to deliver reasonable predictions. Fig. \ref{fig:mask} shows some examples describing that the poses of low-quality pedestrians are much more unstable than the masks estimated using the same Mask RCNN \citep{he2017mask} detector. 
As a result, it is quite infeasible to seamlessly apply these pose transfer models for augmenting current pedestrian datasets.  

In pedestrian detection, some studies have introduced specifically designed GANs for the augmentation. 
As an example, \citep{ouyang2018pedestrian} modified the pix2pixGAN \citep{isola2017image} to make it more suitable for the pedestrian generation, but this method lacks a particular mechanism that helps produce diversified pedestrians and the method still delivers poor generation results based on low-quality data.

In this study, we propose that transforming pedestrians from the original dataset by altering their shapes can produce diversified and much more lifelike pedestrians without requiring rich appearance details for supervision.

\begin{figure*}[t]
\begin{center}
\includegraphics[width=\textwidth,height=0.44\textheight]{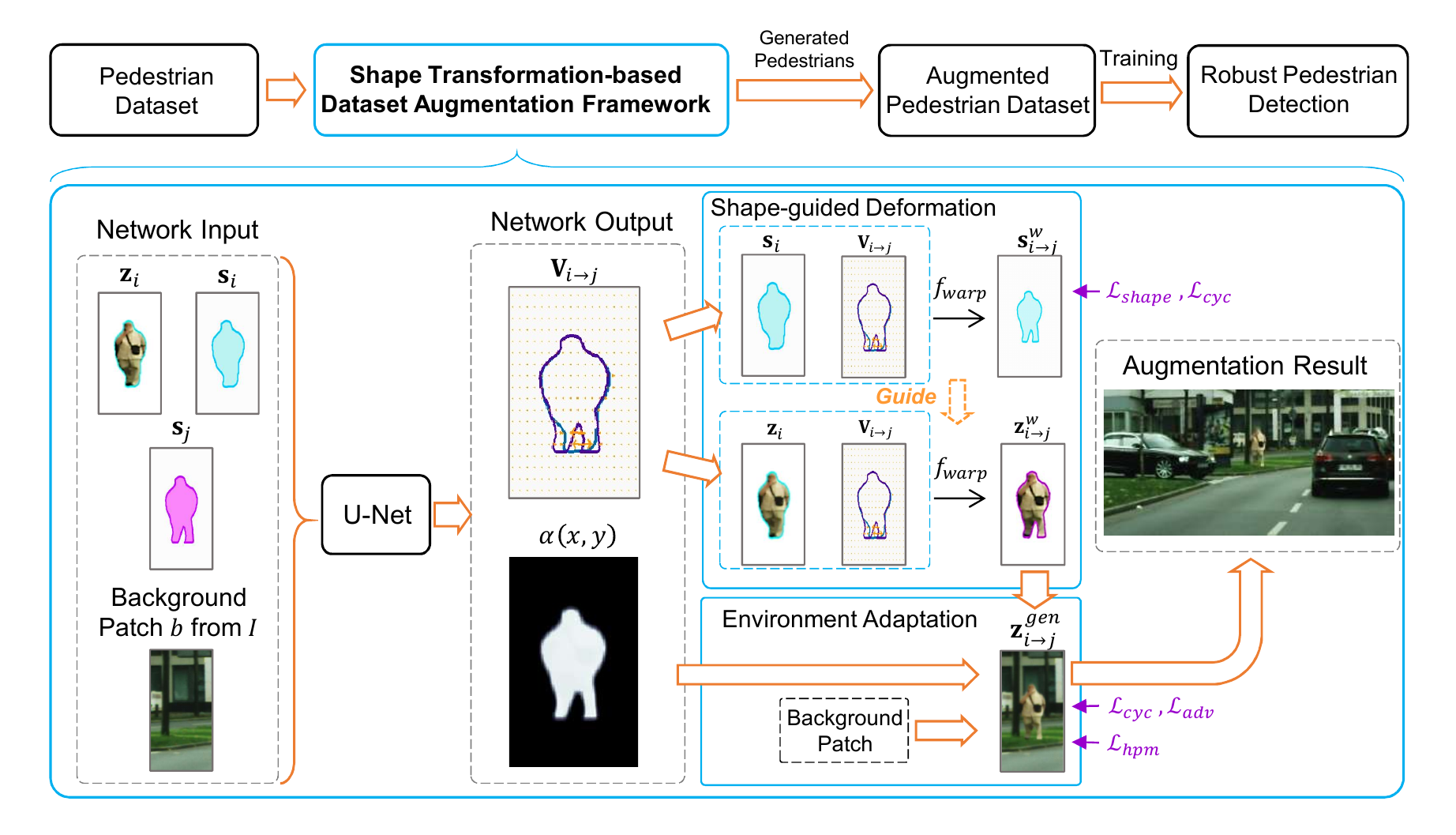}
\end{center}
   \caption{Overview of the proposed shape transformation-based dataset augmentation framework for pedestrian datasets with low-quality pedestrian data. In particular, we introduce the shape-guided warping field, $\mathbf{V}_{i\rightarrow j}$, and the environment-aware blending map, $\mathbf{\alpha}(x,y)$, to respectively help implement the shape-guided deformation and the environment adaptation, obtaining the deformed shape $\mathbf{s}^w_{i\rightarrow j}$, the deformed pedestrian $\mathbf{z}^w_{i\rightarrow j}$, and the transformation result $\mathbf{z}^{gen}_{i\rightarrow j}$. By placing the $\mathbf{z}^{gen}_{i\rightarrow j}$ into the $I$, we can effectively augment the original image. In practice, we employ a U-Net to predict both of the $\mathbf{V}_{i\rightarrow j}$ and $\alpha(x,y)$. Training losses for the U-Net include $\mathcal{L}_{shape}$, $\mathcal{L}_{adv}$, $\mathcal{L}_{cyc}$, and $\mathcal{L}_{hpm}$. Best view in color.}
\label{fig:main}
\end{figure*} 
\section{Shape Transformation-based Dataset Augmentation Framework}

\subsection{Problem Definition}
Data augmentation technique, commonly formulated as transformations of raw data, has been used to access the vast majority of the state-of-the-art results in image recognition. The data augmentation is intuitively explained as to increase the training data size and as a regularizer that can model hypothesis complexity \citep{goodfellow2016deep, zhang2016understanding,dao2019kernel}. In particular, the hypothesis complexity can be used to measure the generalization error, which is the difference between the training and test errors, of learning algorithms \citep{vapnik2013nature,liu2017algorithmic}. Larger hypothesis complexity usually implies a larger generalization error and vice versa. In practice, a small training error and a small generalization error is favoured to guarantee a small test error. As a result, the data augmentation is especially useful for deep learning models which are powerful in maintaining a small training error but has a large hypothesis complexity. 
It has been empirically demonstrated that data augmentation operations can greatly improve the generalization ability of deep models  \citep{cirecsan2010deep, dosovitskiy2015discriminative, sajjadi2016regularization}.

In this study, the overall goal is to devise a more effective dataset augmentation framework to improve pedestrian detection models. The framework is supposed to generate diversified and more realistic-looking pedestrian examples to enrich the corresponding datasets in which real pedestrians are usually in very low-quality. 
We achieve this goal by transforming real pedestrians into different shapes rather than rendering new pedestrians. In fact, the devised data augmentation method focusing on only transforming pedestrians can be efficacious in regularizing the hypothesis complexity. This can be empirically justified by our experiments which show that pedestrian detection performance of the baseline model can be significantly improved without introducing extra model complexity. In general, the top line of Fig. \ref{fig:main} briefly describes our augmentation methodology.

Formally, suppose $\mathbf{z}_i$ is an image patch containing a real pedestrian in the dataset and $\mathbf{s}_i$ is its extracted shape or segmentation mask. Denote $\mathbf{s}_j$ as a different shape which can be obtained based on another real pedestrian's shape.
Here, we implement a shape transformation-based dataset augmentation function, denoted as $f_{STDA}$, to generate a new pedestrian by transforming a real pedestrian into a new pedestrian with a more realistic-looking appearance but with another shape $\mathbf{s}_j$ for the augmentation: 
\begin{equation}
\mathbf{z}^{gen}_{i\rightarrow j} = f_{STDA}(\mathbf{z}_i, \mathbf{s}_j, I),
\end{equation}
where $\mathbf{z}^{gen}_{i\rightarrow j}$ is a patch containing the newly generated pedestrian $\mathbf{z}_i$ by transforming its shape into $\mathbf{s}_j$, and $I$ is the image to be augmented.

\subsection{Framework Overview}
In pedestrian detection datasets, it is difficult to access sufficient appearance details to define the desired $\mathbf{z}^{gen}_{i\rightarrow j}$, making it extremely challenging to generate realistic-looking pedestrians using low-quality appearance. 
To properly implement the $f_{STDA}$ using low-quality pedestrians, we decompose the pedestrian generation task into two sub-tasks, which are shape-guided deformation and environment adaptation. The first task focuses on varying the appearances to enrich data diversity, and the second task mainly adapts the deformed pedestrians different environments to blend the transformed pedestrians within each image to be augmented. 
More specifically, given a pedestrian image patch $\mathbf{z}_i$ and a different shape $\mathbf{s}_j$, we first deform the image $\mathbf{z}_i$ into a pedestrian with similar appearance but a shape $\mathbf{s}_j$ that is different from the original shape $\mathbf{s}_i$. The deformation can be defined according to the transformation from $\mathbf{s}_i$ into $\mathbf{s}_j$. Then, we adapt the deformed pedestrian image into some background environments on the image $I$.
Denote by $f_{SD}$ the function that implements the shape-guided deformation, and 
denote by $f_{EA}$ the function that implements the environment adaptation. The proposed framework implements $f_{STDA}$ as follows:
\begin{equation}
f_{STDA}(\mathbf{z}_i, \mathbf{s}_j, I) = f_{EA}(f_{SD}(\mathbf{z}_i, \mathbf{s}_j),I).
\label{eq:framework}
\end{equation}

Fig. \ref{fig:main} shows a detailed architecture of the proposed framework. As illustrated in the figure, we introduce a shape-guided warping field, denoted as $\mathbf{V}_{i\rightarrow j}$, to help implement the shape-guided deformation function. We also propose to apply the environment-aware blending map, denoted as $\alpha(x,y)$, to achieve environment adaptation. In particular, with the help of $\mathbf{V}_{i\rightarrow j}$, the deformation between different shapes can guide the deformation of appearances of real pedestrians. Then, after better adapting the shape-deformed pedestrian into the background environments using $\alpha(x,y)$, we obtain diversified and more realistic-looking pedestrians to augment pedestrian detection datasets. In practice, we can employ a single end-to-end U-Net \citep{ronneberger2015u} to help fulfill the both sub-tasks in a single pass. According to Eq. \ref{eq:framework}, the employed network takes as input the pedestrian patch $\mathbf{z}_i$, its shape $\mathbf{s}_i$, the target shape $\mathbf{s}_j$, and a background patch from $I$, and then predicts both of the $\mathbf{V}_{i\rightarrow j}$ and the $\alpha(x,y)$.
 
In the following sections, we will subsequently describe in details the implementation of shape-guided deformation in Sec. \ref{sec:SD} and the implementation of environment adaptation in Sec. \ref{sec:EA}.

\subsubsection{Shape-guided Deformation}
\label{sec:SD} 

In this study, we implement deformation according to warping operations. In order to obtain a detailed description about warping operations, we introduce the \textit{shape-guided warping field}, which is the assignment of the vectors for warping between shapes, to further help deform pedestrians.
Denote by $\mathbf{v}_{i\rightarrow j}(x, y)$ the warping vector located at the $(x,y)$ that helps warp the shape $\mathbf{s}_i$ into the shape $\mathbf{s}_j$. The set of these warping vectors, \ie $\mathbf{V}_{i\rightarrow j} = \{\mathbf{v}_{i\rightarrow j}(x, y)\}$, then forms a shape-guided warping field. An example of this warping field can be found in Fig. \ref{fig:main} under the symbol $\mathbf{V}_{i\rightarrow j}$, where the warping field helps deform the $\mathbf{s}_i$ (colored in blue) into the $\mathbf{s}_j$ (colored in purple). 
Then, suppose $f_{warp}$ is the function that warps the input image patch according to the predicted warping field, we then implement the $f_{SD}$ by:
\begin{equation}
\mathbf{z}^w_{i\rightarrow j} = f_{SD}(\mathbf{z}_i, \mathbf{s}_j) = f_{warp}(\mathbf{z}_i; \mathbf{V}_{i\rightarrow j}),
\label{eq:warp}
\end{equation}
where $\mathbf{z}^w_{i\rightarrow j}$ is the warped pedestrian $\mathbf{z}_i$ according to the shape $\mathbf{s}_j$. In practice, we define that each warping vector $\mathbf{v}_{i\rightarrow j}(x, y)$ is a 2D vector which contains the horizontal and vertical displacements between the mapped warping point and the original point located at $(x,y)$. Thus, we can make the employed network directly predict the $\mathbf{V}_{i\rightarrow j}$. 
In addition, we implement the $f_{warp}$ with the help of bilinear interpolation, since the bilinear interpolation can properly back-propagate the gradients obtained from $\mathbf{z}^w_{i\rightarrow j}$ to the $\mathbf{V}_{i\rightarrow j}$, aiding the training of the employed network. For more details about using bilinear interpolation for warping and training, we refer readers to \citep{jaderberg2015spatial,dai2017deformable}.

To make the shape-guided warping field adequately describe the deformation between shapes, we define that the estimated warping field should warp the shape $\mathbf{s}_i$ into the shape $\mathbf{s}_j$. Accordingly, the desired warping field $\mathbf{V}_{i\rightarrow j}$ should make the following equation hold:
\begin{equation}
\mathbf{s}_j = \mathbf{s}^w_{i\rightarrow j} = f_{warp} ( \mathbf{s}_{i};\mathbf{V}_{i\rightarrow j}),
\label{eq:shape}
\end{equation}
where $\mathbf{s}^w_{i\rightarrow j}$ is the warped shape $\mathbf{s}_{i}$ according to $\mathbf{V}_{i\rightarrow j}$.
Since $\mathbf{s}_i$ and $\mathbf{s}_j$ can be easily obtained from the pedestrian datasets, we are able to access sufficient pixel-level supervision to train the employed network. We mainly apply the $L_1$ loss, $||\mathbf{s}_j - \mathbf{s}^w_{i\rightarrow j}||_1$, to make the network satisfy the above definition during the training. Therefore, by satisfying the above definition, we can obtain the desired warping field that can help generate shape-transformed natural pedestrians based on Eq. \ref{eq:warp}.

\emph{Shape Constraining Operation:}
In practice, we observe that if the target shape $\mathbf{s}_j$ varies too much \wrt $\mathbf{s}_i$, the obtained warping field may distort the input pedestrians after warping, resulting in unnatural results that could degrade the augmentation performance. To avoid this, we apply a \textit{shape constraining operation} on the target shape. 

\begin{figure}[t]
\begin{center}
\includegraphics[width=\linewidth,height=0.18\textheight]{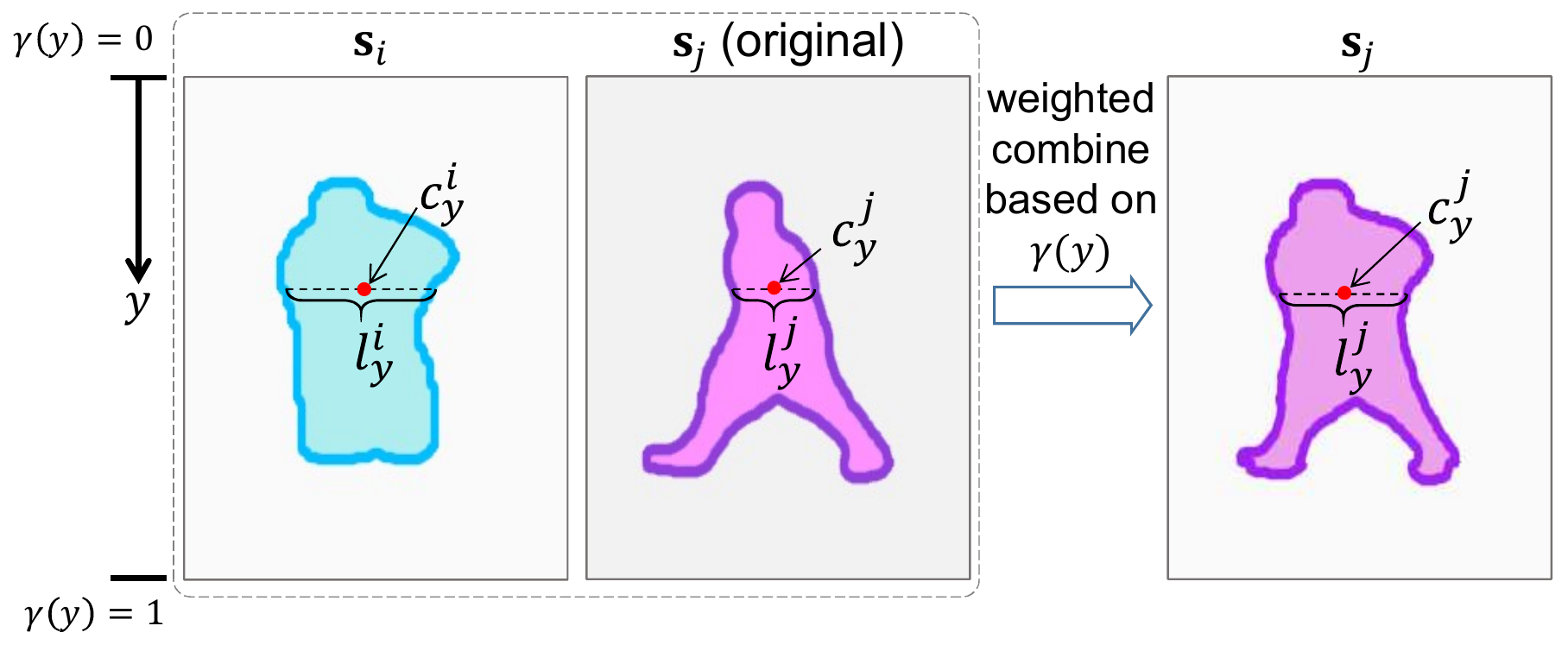}
\end{center}
   \caption{An example of the shape constraining operation. In particular, we combine the shape $\mathbf{s}_j$ with $\mathbf{s}_i$ based on a weighting function $\gamma(y)$. $c^j_y$ and $l^j_y$ respectively denote the middle point and the width of the foreground areas on the line whose vertical offset is $y$ and on the shape $\mathbf{s}_j$. $c^i_y$ and $l^i_y$ are the corresponding middle point and the width of foreground areas on the shape $\mathbf{s}_i$. The $\gamma(y)$ is a linear function of $y$, controlling the combination of $\mathbf{s}_i$ and $\mathbf{s}_j$. Best view in color.}
\label{fig:cstr}
\end{figure}
To avoid inadequate shape deformation, we hypothesize that varying more for the lower body of a natural pedestrian is more acceptable than varying more for the upper body. In particular, we find that changing too much the upper body of a pedestrian would generally require the change of the viewpoint for that pedestrian (\eg from the side view to the front view) to obtain a natural appearance, but the warping operations do not generate new image contents to generate the pedestrian in a different viewpoint.
As a result, we apply a constraint for the target shape $\mathbf{s}_j$ to avoid varying too much for the upper body.
More specifically, we constrain the target shape $\mathbf{s}_j$ by combining it with the input shape $\mathbf{s}_i$ according to a weighted function.
The combination is defined on the middle point and the width of the foreground areas in each horizontal line on the $\mathbf{s}_j$. Suppose $y$ is the vertical offset for a horizontal line on $\mathbf{s}_j$. We respectively denote $c^j_y$ and $l^j_y$ as the middle point and the width of the foreground areas on that line. We define the shape constraining operation as:
\begin{equation}
\left\{
\begin{array}{ll}
c^j_y &= \gamma(y) ~ c^j_y + (1-\gamma(y)) ~ c^i_y, \\
l^j_y &= \gamma(y) ~l^j_y + (1-\gamma(y))~ l^i_y, 
\end{array}
\right.
\label{eq:cstr}
\end{equation}
where $c^i_y$ and $l^i_y$ refer to the middle point and width of foreground areas on the $\mathbf{s}_i$ at line $y$, and $\gamma(y)$ is the weight function \wrt $y$ controlling the strictness of the constraint. According to Eq. \ref{eq:cstr}, the smaller weight value $\gamma(y)$ can make the target shape contribute less to the combination result and \textit{vice versa}. 
Accordingly, to allow more variations only for the lower body of a pedestrian, we define $\gamma(y)$ as a linear function which increases from 0 to 1 with $y$ varying from the top to the bottom. 

Fig. \ref{fig:cstr} shows a visual example of the introduced shape constraining operation when constraining $\mathbf{s}_j$ according to Eq. \ref{eq:cstr}. We can observe that the proposed shape constraining operation adequately constrains the $\mathbf{s}_j$ by making the output shape closer to $\mathbf{s}_i$ for the upper body of the pedestrian and closer to $\mathbf{s}_j$ for the lower body.

\subsubsection{Environment Adaptation}
\label{sec:EA}
After the shape-guided deformation, we place the deformed pedestrians into the image $I$ to fulfill augmentation. 
However, directly pasting deformed pedestrians on the image would sometimes produce significant appearance mismatch due to the issues like the discontinuities in illumination conditions and imperfect shapes predicted by the mask extractor. To refine the generated pedestrians according to the environments, we further perform environment adaptation. 

To properly blend a shape-deformed pedestrian into the image $I$ by considering surrounding environments, we introduce an environment-aware blending map to help refine the deformed pedestrians. We formulate this refinement procedure as follows:
\begin{equation}
\mathbf{z}^{gen}_{i\rightarrow j} = f_{EA}(\mathbf{z}^w_{i\rightarrow j}  , I) = 
\{\mathbf{z}^{a}_{i\rightarrow j}(x,y)\}, 
\label{eq:ea}
\end{equation}
where $\mathbf{z}^{a}_{i\rightarrow j}(x,y)$ is the environment adaptation result located at $(x,y)$:
\begin{multline}
\mathbf{z}^{a}_{i\rightarrow j}(x,y)
= \Big(\mathbf{s}_j(x,y)  \cdot \mathbf{\alpha}(x,y)\Big) \cdot \mathbf{z}^w_{i\rightarrow j}(x,y) \\ + \Big(1-\mathbf{s}_j(x,y)  \cdot \mathbf{\alpha}(x,y))\Big) \cdot I(x,y),
\label{eq:ea-2}
\end{multline}
where $\mathbf{\alpha}(x,y)$ is an entry value of the environment-aware blending map located at $(x,y)$. An example of the estimated  $\mathbf{\alpha}(x,y)$ can be found in Fig. \ref{fig:main}.

In practice, it is difficult to define the desired refinement result and the desired environment-aware blending map. Therefore, we can not access appropriate supervision information to train the employed network for environment adaptation. Without supervision, we apply an adversarial loss to facilitate the employed network to learn and blend the deformed pedestrians into the environments effectively. Similar to the shape-guided warping field, we make the employed network directly predict environment-aware blending map. Note that we constrain the environment-aware blending map to prevent changing the appearance of the deformed pedestrians too much. In particular, we adopt a shifted and rescaled \textit{tanh} squashing function to make the values of $\alpha(x,y)$ lie in a range of 0.8 and 1.2.

\subsection{Objectives}
Since we employ a single network to predict both the shape-guided warping field and the environment blending map, we can unify the objectives for training.

First, to obtain a proper shape-guided warping field, we introduce a shape deformation loss and a cyclic reconstruction loss. The shape deformation loss ensures that the predicted warping field satisfies the constrain as described in Eq. \ref{eq:shape}. The cyclic reconstruction loss then ensures that the deformed shape and pedestrian can be deformed back to the input shape and pedestrian. Therefore, we define that the shape deformation loss function $\mathcal{L}_{shape}$ for a pair of samples $i \rightarrow j$ is as follows:
\begin{equation}
\mathcal{L}_{shape} = \mathbb{E}[||\mathbf{s}_j - \mathbf{s}^{w}_{i\rightarrow j}||_1 ],
\label{eq:shape_loss}
\end{equation}
and the cyclic loss is defined as follows:
\begin{equation}
\mathcal{L}_{cyc} = \mathbb{E}[ ||\mathbf{s}_i - \mathbf{s}^{w}_{j\rightarrow i}||_1 + ||\mathbf{z}_i - \mathbf{z}^{w}_{j\rightarrow i}||_1],
\label{eq:cyc_loss}
\end{equation}
where $\mathbf{s}^w_{j\rightarrow i}$ is the deformation result of $\mathbf{s}^w_{i\rightarrow j}$ according to $\mathbf{s}_i$ and $\mathbf{z}^w_{j\rightarrow i}$ is the deformation result of $\mathbf{z}^w_{i\rightarrow j}$ using the same warping field for computing $\mathbf{s}^w_{j\rightarrow i}$. As a result, the Eq. \ref{eq:shape_loss} describes the $L_1$-based shape deformation loss, and the Eq. \ref{eq:cyc_loss} form the cyclic reconstruction loss.

In addition, an adversarial loss, denoted as $\mathcal{L}_{adv}$, is included to make sure that the shape-guided deformation and environment adaptation can help produce more realistic-looking pedestrian patches. Similar with typical GANs, the adversarial loss is computed by introducing a discriminator $D$ for the employed network:
\begin{equation}
\mathcal{L}_{adv}= \mathbb{E}[\log D(\mathbf{z})]
 +\mathbb{E}[\log\left(1-D(\mathbf{z}^{gen}_{i\rightarrow j})\right)],
\label{eq:adv}
\end{equation}
where $\mathbf{z}$ refers to any real pedestrian in the dataset.

\emph{Hard Positive Mining Loss:} 
Since our final goal is to improve the detection performance, we further apply a hard positive mining loss to magnify the benefits of the transformed pedestrians on improving detection robustness. Inspired by the study of hard positive generation \citep{wang2017fast}, we attempt to generate pedestrians that are not very easy to be recognized by a RCNN detector \citep{girshick2014rich}. Different from the study \citep{wang2017fast} that additionally introduced an occlusion mask and the spatial transformation operations to generate hard positives, we only introduce a loss function on the transformed pedestrians to make the employed network learn to produce harder positives for the RCNN detector. To compute this loss, we additionally train a RCNN, denoted as $R$, to distinguish pedestrian patches from background patches which do not contain pedestrians inside. Suppose $\mathcal{L}_{hpm}$ is the hard positive mining loss, then we have:
\begin{equation}
\mathcal{L}_{hpm} =\mathbb{E}[\log(1-R(\mathbf{z}^{gen}_{i\rightarrow j}))] + \mathbb{E}[\log R(\mathbf{z})] +\mathbb{E}[\log(1-R(\mathbf{b}))],
\label{eq:m}
\end{equation}
where $\mathbf{b}$ refers to background image patches in the dataset. 

The major difference between the $\mathcal{L}_{hpm}$ and the $\mathcal{L}_{adv}$ is that the $R$ distinguishes between pedestrians patches and background patches, while $D$ in $\mathcal{L}_{adv}$ distinguishes between true pedestrian patches and the shape-transformed pedestrian patches.

\textbf{Overall Loss.} 
To sum up, the overall training objective $\mathcal{L}$ of the network employed to help implement the proposed framework can be written as follows:
\begin{equation}
\mathcal{L} = \omega_1 \mathcal{L}_{shape} +  \omega_2\mathcal{L}_{cyc} +  \omega_3\mathcal{L}_{adv} +  \omega_4\mathcal{L}_{hpm},
\end{equation}
where $\omega_1$, $\omega_2$, $\omega_3$, $\omega_4$ are the corresponding loss weights. In general, we borrow the setting from the implementation of pix2pixGAN\footnote{https://github.com/junyanz/pytorch-CycleGAN-and-pix2pix} and set the $\omega_1$ and $\omega_3$ to 100 and 1, respectively. Since we find in the experiment that the network can hardly learn a proper shape-guided warping field if $\omega_2$ is too large, we empirically set the $\omega_2$ to a small value, \ie ~0.5, in this study.
Similarly, we also set $\omega_4$ to 0.5 to make the hard positive mining loss contribute less to the overall objective. In practice, the network is obtained by minimizing the overall loss $\mathcal{L}$, the discriminator $D$ is obtained by maximizing the $\mathcal{L}_{adv}$, and the $R$ is obtained by maximizing the $\mathcal{L}_{hpm}$. 

\subsection{Dataset Augmentation} 
When augmenting the pedestrian datasets with the proposed framework, we attempted to sample more natural locations and sizes to place the transformed pedestrians in the image. 
Fortunately, pedestrian datasets deliver sufficient knowledge encoded within bounding box annotations to define these geometric statistics of a natural pedestrian.
For example, in the Caltech dataset \citep{dollar2009pedestrian,dollar2012pedestrian}, the aspect ratio of a pedestrian is usually around 0.41. In addition, it is also possible to describe the bottom edge $y^{box}$ and the height $h^{box}$ of an annotated bounding box for a pedestrian using a linear model \citep{park2010multiresolution}: $h^{box} = k y^{box} + b$,
where $k$ and $b$ are the coefficients. In the Caltech dataset whose images are 480 by 640, the $k$ and $b$ are found to be around 1.15 and -194.24. 
For each image to be augmented, we sample several locations and sizes according to this linear model. Then, for each sampled location and size, we run the proposed framework and put the transformation result into the image. 
Algorithm \ref{alg:aug} describes the detailed pipeline of applying the proposed framework to augment pedestrians dataset. 

\begin{algorithm}[h]   
\caption{Pedestrian Dataset Augmentation Pipeline}   
\label{alg:aug}   
\begin{algorithmic}[1] 
\REQUIRE  
Natural pedestrians $\{\mathbf{z}_i\}$, the corresponding shapes $\{\mathbf{s}_i\}$, and images from original dataset $\mathcal{I}=\{I\}$. 
\ENSURE  
Augmented dataset images $\mathcal{I}$;\\  
\FOR{each image $I \in \mathcal{I}$}
\STATE Uniformly sample a number $n$ from the set $\{1, 2, 3, 4, 5\}$;
\FOR{$m = 1: n$}
\STATE Sample a location and a size according to $h^{box} = k y^{box} + b$; 
\STATE Sample a pedestrian patch $\mathbf{z}_i$, a shape $\mathbf{s}_j$ from a different pedestrian, and a background patch cropped from the sampled location and size on $I$ ; 
\STATE Perform shape-guided deformation according to Eq. \ref{eq:warp} and then perform environment adaptation according to Eq. \ref{eq:ea}, obtaining $\mathbf{z}^{gen}_{i\rightarrow j}$;
\STATE Place the $\mathbf{z}^{gen}_{i\rightarrow j}$ into the image $I$ according to the sampled location and size;
\ENDFOR
\ENDFOR
\RETURN $\mathcal{I}$
\end{algorithmic}  
\end{algorithm}

\section{Experiments}
We perform comprehensive evaluation for the proposed STDA framework to augment pedestrian datasets. 
We use the popular Caltech \citep{dollar2009pedestrian,dollar2012pedestrian} and CityPersons \citep{zhang2017citypersons} benchmarks for the evaluation. 

In this section, we will first present the overall dataset augmentation results on evaluated datasets. Then, we will validate the improvements in improving detection accuracy of applying our proposed STDA framework to augment different datasets, comparing to other cutting-edge pedestrian detectors. Subsequently, we perform detailed ablation studies on the STDA framework to analyze the effects of different components in STDA on generating more realistic-looking pedestrians and on improving detection accuracy.

\subsection{Settings and Implementation Details}
\label{sec:impl}
For evaluation, we consider the log-average miss rates (MR) against different false positive rates as the major metric to represent pedestrian detection performance. 
In the Caltech, we follow the protocol of \citep{zhang2016faster} and use around 42k images for training and 4024 images for testing. In the Citypersons, as suggested in the original study, we use 2975 images for training and perform the evaluation on the 500 images from the validation set. We apply a Mask RCNN to extract shapes on Caltech and use the annotated pedestrian masks on Citypersons. 
To augment the datasets, for each frame, we transform $n$ pedestrians using our framework and $n$ is uniformly sampled from $\{1, 2, 3, 4, 5\}$.
Thus, each image has the number of positive pedestrians increased by 1 $\sim$ 5. 

For the network employed to implement the framework, we use the U-net architecture with 8 blocks. 
All the input and output patches have a size of 256 $\times$ 256. Both the $D$ as introduced in Eq. \ref{eq:adv} and the $R$ as introduced in Eq. \ref{eq:m} are CNNs with 3 convolutional blocks. During optimization, we reduce the updating frequency of $D$ and $R$ to stabilize the training, \textit{i.e.} we update $D$ and $R$ once at every 40-th update of the U-net. Learning rate is set to $1e-5$ and we perform training with 80 epochs for a dataset. 

We adopt a ResNet50-based FPN detector \citep{lin2017feature} as our baseline detector. When training this detector, we modified some default parameters according to the pedestrian detection task. First, for the region proposal network in FPN, we follow the \citep{zhang2016faster} and only use the anchor with the aspect ratio of 2.44. We discard the 512x512 anchors in FPN because they do not contribute much to the performance. In addition, we set the batch size as 512 for both the Region Proposal Network (RPN) and the Regional-CNN (RCNN) in FPN. To reduce of false positive rates of FPN, we further set the foreground thresholds of RPN and RCNN to 0.5 and 0.7, respectively. During training, we make the length of the shorter size of input images as 720 for Caltech and as 1024 for CityPersons. We train the FPN detector on the Caltech with 3 epochs and on the CityPersons with 6 epochs. In general, the final performance of the baseline detector is $10.4\%$ mean miss rate on Caltech test set and $13.9 \%$ mean miss rate on CityerPersons validation set. Note that we weight the loss values for synthesized pedestrians by a factor of 0.1, reducing the potential biases towards generated pedestrians rather than real pedestrians.

\begin{figure*}[t]
\begin{center}
\includegraphics[width=\linewidth,height=0.78\textheight]{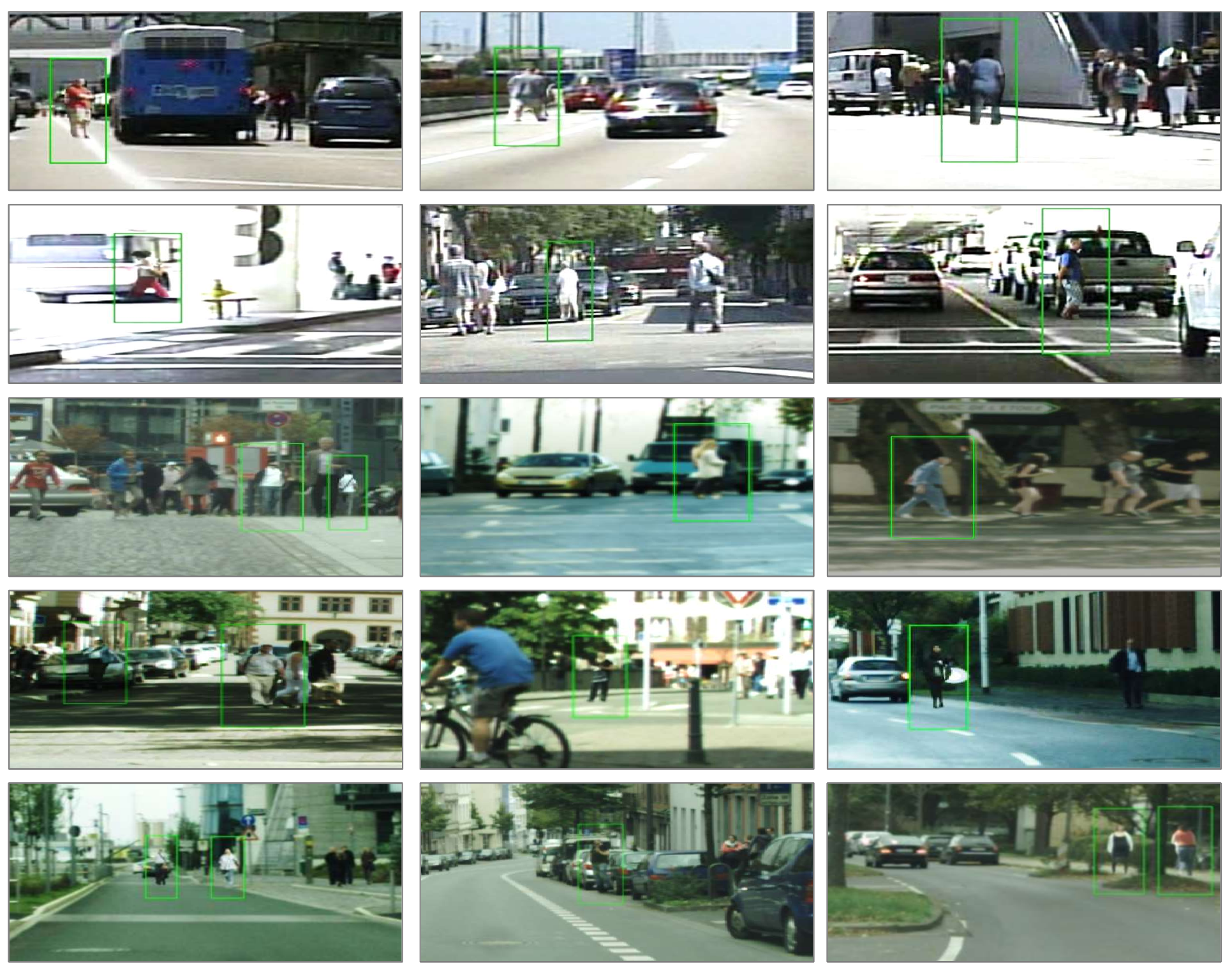}
\end{center}
   \caption{Dataset augmentation results of the proposed STDA on images from Caltech (top 2 rows) and CityPersons (bottom 3 rows), respectively. Light green bounding boxes indicate the synthesized pedestrians. The presented image patches are cropped and zoomed to better illustrate the details. Best view in color.}
\label{fig:vis_gen}
\end{figure*}

\begin{figure}[h]
    \centering
    \subfigure[Synthesis results of PS-GAN \citep{ouyang2018pedestrian}.\label{fig:cmp-gans-a}]{
        \centering
        \includegraphics[width=0.45\textwidth,height=0.3\textheight]{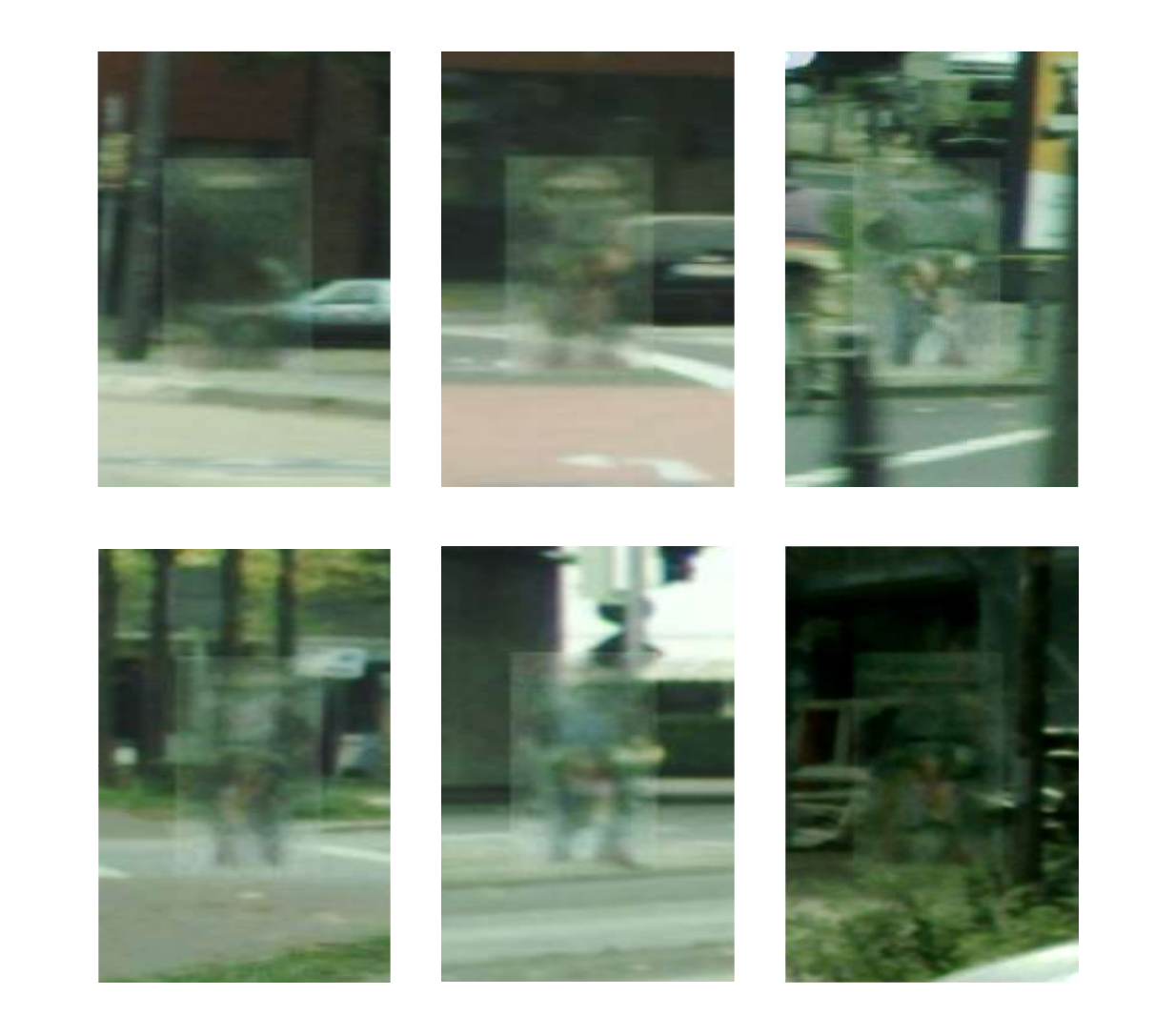}
    }
    \subfigure[Synthesis results of STDA (ours)\label{fig:cmp-gans-b}]{
        \centering
        \includegraphics[width=0.45\textwidth,height=0.3\textheight]{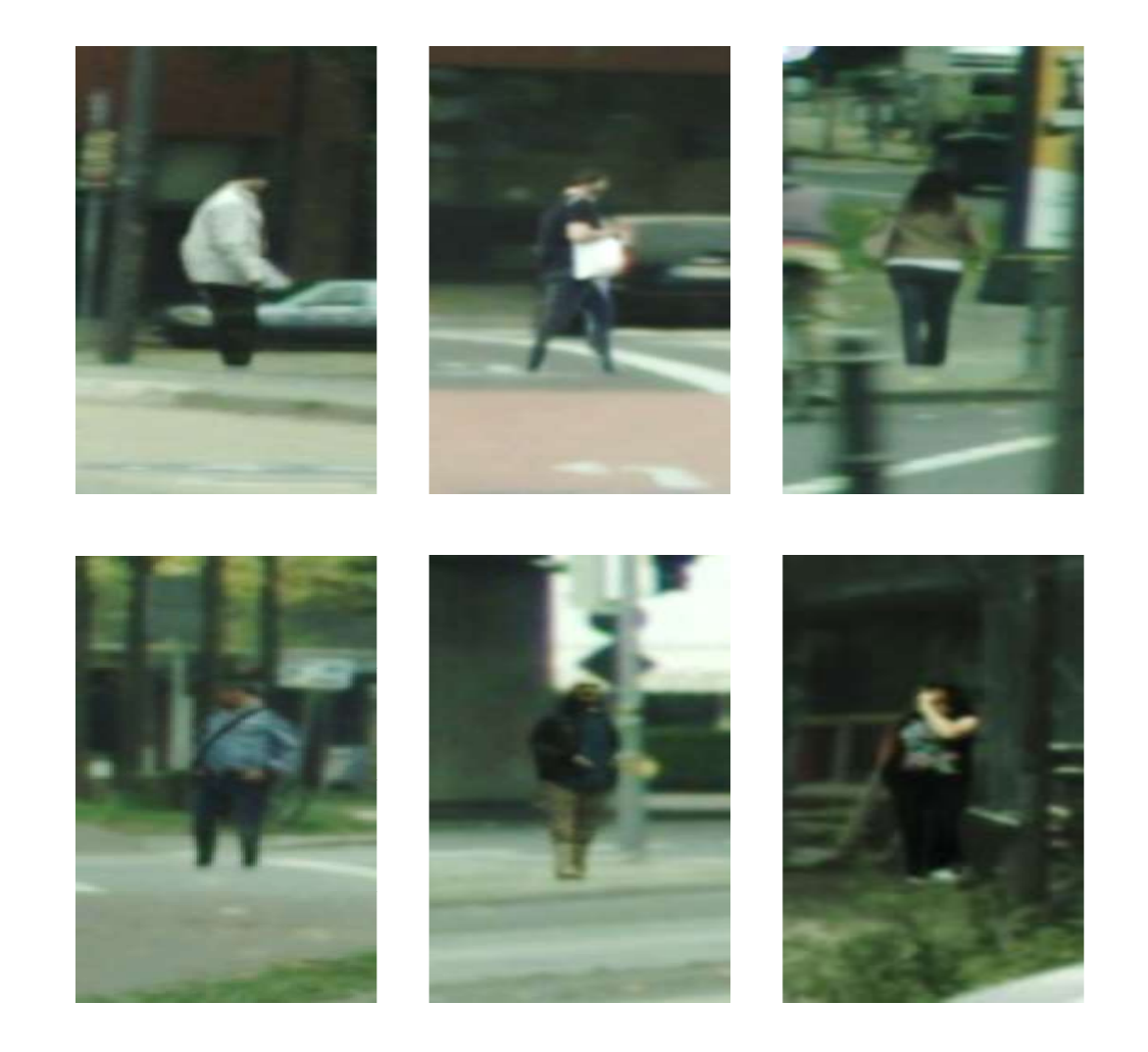}
     }
    \caption{Pedestrian synthesis results of STDA, comparing to another cutting-edge GAN-based data generation method. Synthesized pedestrians are in the middle of each image patch and background patches are kept the same. Best view in color.}
\end{figure}
\subsection{Dataset Augmentation Results}
We first present the pedestrian synthesis results of applying our STDA framework to augment pedestrian datasets. 

\subsubsection{Pedestrian Synthesis Results}
In Figure \ref{fig:vis_gen}, we illustrate the dataset augmentation results on both the evaluated Caltech dataset and citypersons dataset. Even if some of the pedestrians are blurry and lack rich appearance details, we can observe that the shape transformed pedestrians can still be naturally blended into the environments of the image, obtaining very realistic-looking pedestrians for dataset augmentation. Furthermore, the STDA can also produce pedestrians in uncommon walking areas, such as in the middle areas of the street. This can increase the irregular foreground examples for pedestrian detection, and the model can be more robust in detecting pedestrians after augmentation. Moreover, with a similar geometry arrangement with real pedestrians, the illustrated results can demonstrate that the proposed STDA framework is effective in generating pedestrians in a similar domain with real pedestrians. Besides, our method can produce occlusion cases, \textit{e.g.} by overlapping the generated pedestrians over real pedestrians, which can promisingly increase the amount of occlusion cases for training and thus improve the detection robustness for occlusions.

In addition, we also compare our method with another recently published powerful GAN-based data rendering technique, \ie~PS-GAN \citep{ouyang2018pedestrian}, using the same background patches. We train the compared PS-GANs using similar protocols with the original study, while the major difference is that we use very \emph{low-quality} pedestrian data provided in the pedestrian dataset. Training schemes for both our study and the PS-GAN are kept the same. Figure \ref{fig:cmp-gans-a} shows some pedestrian synthesis results using existing GANs. We can find that the compared GAN-based method produces very blurry pedestrians due to low-quality training data. Besides, the generated backgrounds can be also unnatural and distorted due to the lack of sufficient appearance details during training. On the contrary, as shown in Figure \ref{fig:cmp-gans-b}, our proposed STDA framework can effectively generate much more realistic and natural-looking pedestrians in different background patches. 
our method achieves significantly lower score than PS-GAN, meaning that the STDA-generated pedestrians are much more similar to the true data. 
This illustrates its superiority over the GAN-based data rendering methods. 

\subsubsection{Improvements for Pedestrian Detection}
\label{sec:cal}  

\begin{figure}[h]
\begin{center}
\includegraphics[width=\linewidth]{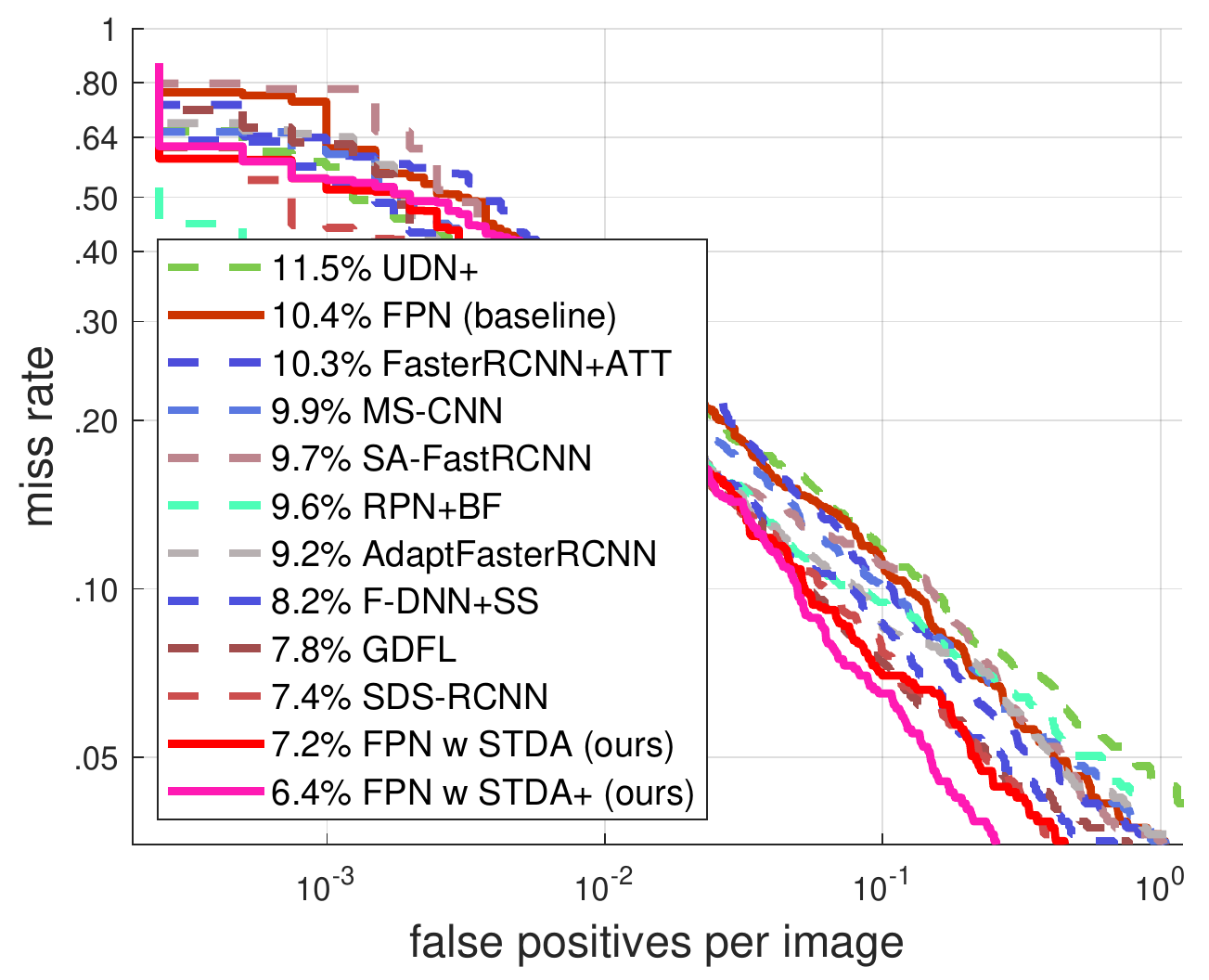}
\end{center}
   \caption{Effects of the proposed STDA framework for augmenting the Caltech pedestrian dataset, comparing to other cutting-edge pedestrian detectors. ``+'' means multi-scale testing.}
\label{fig:Caltech}
\end{figure}

\begin{figure}[h]
\begin{center}
\includegraphics[width=\linewidth]{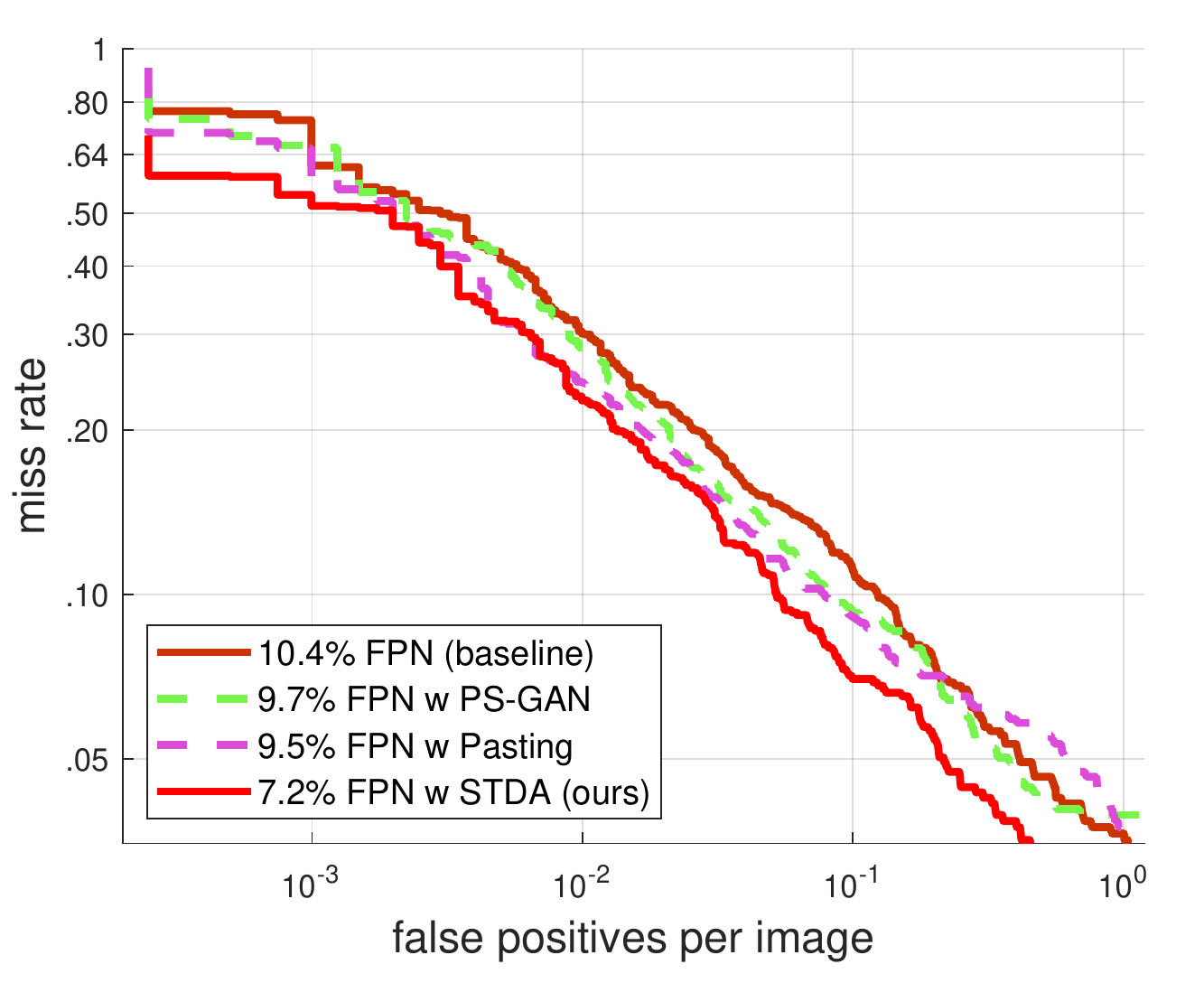}
\end{center}
\vspace{-0.2cm}
   \caption{Effects of the proposed STDA framework for augmenting the Caltech pedestrian dataset, comparing to other pedestrian synthesis methods such as random pasting, PoseGen \citep{ma2017pose}, and PS-GAN \citep{ouyang2018pedestrian}. }
\label{fig:CaltechGAN}
\end{figure}
\begin{table*}[h]
\begin{center}
\resizebox{0.7\linewidth}{!}{
  \begin{tabular}{|c |c c|}
  \hline
 Methods/MR(\%) & OCC (heavy) & OCC (partial) \\ %
  \hline
  UDN+ \citep{ouyang2018jointly}&70.3&28.2\\ %
  FasterRCNN+ATT \citep{zhang2018occluded} &45.2& 22.3\\ %
  MS-CNN \citep{cai2016unified}  &59.9& 19.2\\ %
  SA-FastRCNN \citep{li2018scale}&64.4 & 24.8\\ %
  RPN+BF \citep{zhang2016faster} &74.4 & 24.2\\ %
  AdaptFasterRCNN \citep{zhang2017citypersons}  &57.6& 26.5\\ %
  F-DNN+SS \citep{du2017fused} &53.8&15.1\\ %
  GDFL \citep{lin2018graininess} &43.2 &16.7\\ %
  SDS-RCNN \citep{brazil2017illuminating}  &58.5& 14.9 \\ %
  FPN (baseline) \citep{lin2017feature}  &59.3 &22.9\\ %
  \hline
  FPN w STDA (ours)&\textbf{41.9}&17.2 \\ %
  FPN w STDA+ (ours)&43.5 &\textbf{12.4}\\ %
  \hline
  \end{tabular}
}
\end{center}
\caption{Performance on occluded (OCC) pedestrians on Caltech test set. Best results are highlighted in \textbf{bold}. ``+'' means multi-scale testing.}
\label{tab:occ}
\end{table*}

\begin{table*}[h]
\begin{center}
\resizebox{0.7\linewidth}{!}{
  \begin{tabular}{|c |c c |}
  \hline
 Methods/MR(\%) & AR(typical)& AR(a-typical)\\%
  \hline
  UDN+ \citep{ouyang2018jointly}&7.8&14.9\\%
  FasterRCNN+ATT \citep{zhang2018occluded} &6.0 &19.4 \\%
  MS-CNN \citep{cai2016unified}  &6.3&15.7\\%
  SA-FastRCNN \citep{li2018scale}& 5.7&15.8 \\%
  RPN+BF \citep{zhang2016faster} &6.0&14.5 \\%
  AdaptFasterRCNN \citep{zhang2017citypersons}  &5.0 &16.2\\%
  F-DNN+SS \citep{du2017fused} & 5.1&13.3 \\%
  GDFL \citep{lin2018graininess} &4.5 &14.7 \\%
  SDS-RCNN \citep{brazil2017illuminating} &4.6 &11.7\\%
  FPN (baseline) \citep{lin2017feature} &6.6&15.7 \\%
  \hline
  FPN w STDA (ours)&4.5&11.3 \\%
  FPN w STDA+ (ours)&\textbf{3.9}&\textbf{9.9} \\%
  \hline
  \end{tabular}
}
\end{center}
\caption{Performance on pedestrians with diversified aspect ratios (AR) on Caltech test set. Best results are highlighted in \textbf{bold}. ``typical'' means the pedestrians with normal aspect ratios; ``a-typical'' means the pedestrians with unusual aspect ratios; ``+'' means multi-scale testing.}
\label{tab:ar}
\end{table*}

\textbf{Caltech:}
To evaluate the augmentation results of our proposed STDA framework, we first perform the evaluation on the test set of the Caltech benchmark. We evaluate the performance gains with respect to the baseline detector to demonstrate the effectiveness.
Fig. \ref{fig:Caltech} shows the detailed performance of our method comparing to other cutting-edge methods \citep{ouyang2018jointly,zhang2018occluded,cai2016unified,li2018scale, zhang2016faster,zhang2017citypersons,du2017fused,brazil2017illuminating, lin2018graininess}. 
In particular, our framework improves around 30$\%$ miss rate over the baseline detector. By further applying the multi-scale testing, we can achieve 38$\%$ improvement, significantly out-performing other cutting-edge pedestrian detectors such as the MS-CNN \citep{cai2016unified} that introduced more complicated architecture. 

In Figure \ref{fig:CaltechGAN}, we also compare our framework with some other augmentation methods, including Pasting that directly pastes real pedestrians randomly and PS-GAN \citep{ouyang2018pedestrian} that generates pedestrian patches based on a pix2pixGAN \citep{isola2017image} pipeline. 
We can find that the three other compared methods can also slightly improve the baseline detector, suggesting that augmenting pedestrian datasets with synthesized pedestrians is useful for improving detection accuracy. However, due to the unnatural pedestrians synthesized based on low-quality data as presented in Figure \ref{fig:cmp-gans-a}, improvements brought by PS-GAN is very limited. Even random pasting real pedestrians can deliver a slightly better improvements using low-quality data.
Moreover, we can also observe that the compared methods have higher false positives per image than the baseline detector at low miss rates, suggesting that the baseline detector may be distracted by unnatural pedestrians to some extents. Comparing to the other compared pedestrian synthesis methods, the performance gain brought by our proposed STDA is much more significant with respect to the baseline detector, confirming that our proposed framework is much more effective in augmenting pedestrian datasets using low-quality pedestrian data. Furthermore, with the more realistic-looking pedestrians synthesized by STDA, the augmented dataset can consistently improve the baseline detector at all presented false positives per image.

\begin{table*}[h]
\begin{center}
\resizebox{0.75\linewidth}{!}{
  \begin{tabular}{c c c c c}
  \hline
 Method/MR\% & Reasonable & Heavy & Partial & Bare \\
  \hline
  Citypersons \citep{zhang2017citypersons} & 15.4 &-& - & - \\ %
  TLL \citep{Song_2018_ECCV} & 14.4 &52.0& 15.9& 9.2\\
  RepultionLoss \citep{wang2018repulsion} & 13.2 &56.9 &16.8 &7.6\\
  OR-CNN \citep{zhang2018occlusion} & 12.8 &55.7 &15.3& 6.7\\
  FPN (baseline) \citep{lin2017feature}&13.9 & 52.9& 15.4 & 8.5\\
  \hline
  FPN w STDA (ours) &{11.0}& {44.1} & {11.3}& {6.4}\\
  FPN w STDA+ (ours) &\textbf{10.2}& \textbf{41.9} & \textbf{10.5}& \textbf{5.8}\\
  \hline
  \end{tabular}
}
\end{center}
\caption{Performance on the validation set of CityPersons. Best results are highlighted in \textbf{bold}. ``+' means multi-scale testing.}
\label{tab:citypersons}
\end{table*}

Besides overall performance, we also present performance on specific detection attributes. For example, Table \ref{tab:occ} shows the detection accuracy on pedestrians with partial or heavy occlusions. According to the statistics, we can find that the proposed STDA can effectively reduce the average miss rate of the baseline detector for both partial and heavy occluded pedestrians, achieving favorable performance comparing to other cutting-edge pedestrian detectors. This confirms that synthesizing pedestrians with occlusions using our proposed STDA framework can promisingly help improve the detection robustness and accuracy of occluded pedestrians in test set. In addition, we also evaluate the performance of applying STDA to augment the Caltech on pedestrians with different aspect ratios in the Table \ref{tab:ar}.
In particular, for the detection on the pedestrians with ``typical'' aspect ratios, our proposed framework is able to boost the performance of the baseline detector by up to 41\%. When detecting the pedestrians with ``a-typical'' aspect ratios, our method also promisingly improves the baseline performance, obtaining the highest average miss rate among compared pedestrian detectors. These results demonstrate that our framework can produce rich diversified and beneficial pedestrians for the augmentation.

\textbf{CityPersons:}
\label{sec:cit}  
In this section, we also report the performance on the validation set of CityPersons. The experiment settings are similar to the evaluation for the Caltech dataset except that image sizes are 1024 $\times$ 2048 for training and testing. 

Table \ref{tab:citypersons} presents the detailed statics of the evaluated methods. We can find that our framework effectively augments the original dataset and improves the performance of the baseline FPN detector. State-of-the-art performance can be accessed by applying our proposed framework, further demonstrating that our proposed framework can consistently augment different pedestrian datasets with low-quality pedestrian data.

\subsection{Ablation Studies}
In this section, we perform comprehensive component analysis of the proposed STDA framework for both the pedestrian generation and the pedestrian detection augmentation, using the low-quality pedestrians in Caltech dataset and the Caltech benchmark for training. 

\begin{figure*}[t]
\begin{center}
\includegraphics[width=0.8\linewidth,height=0.5\textheight]{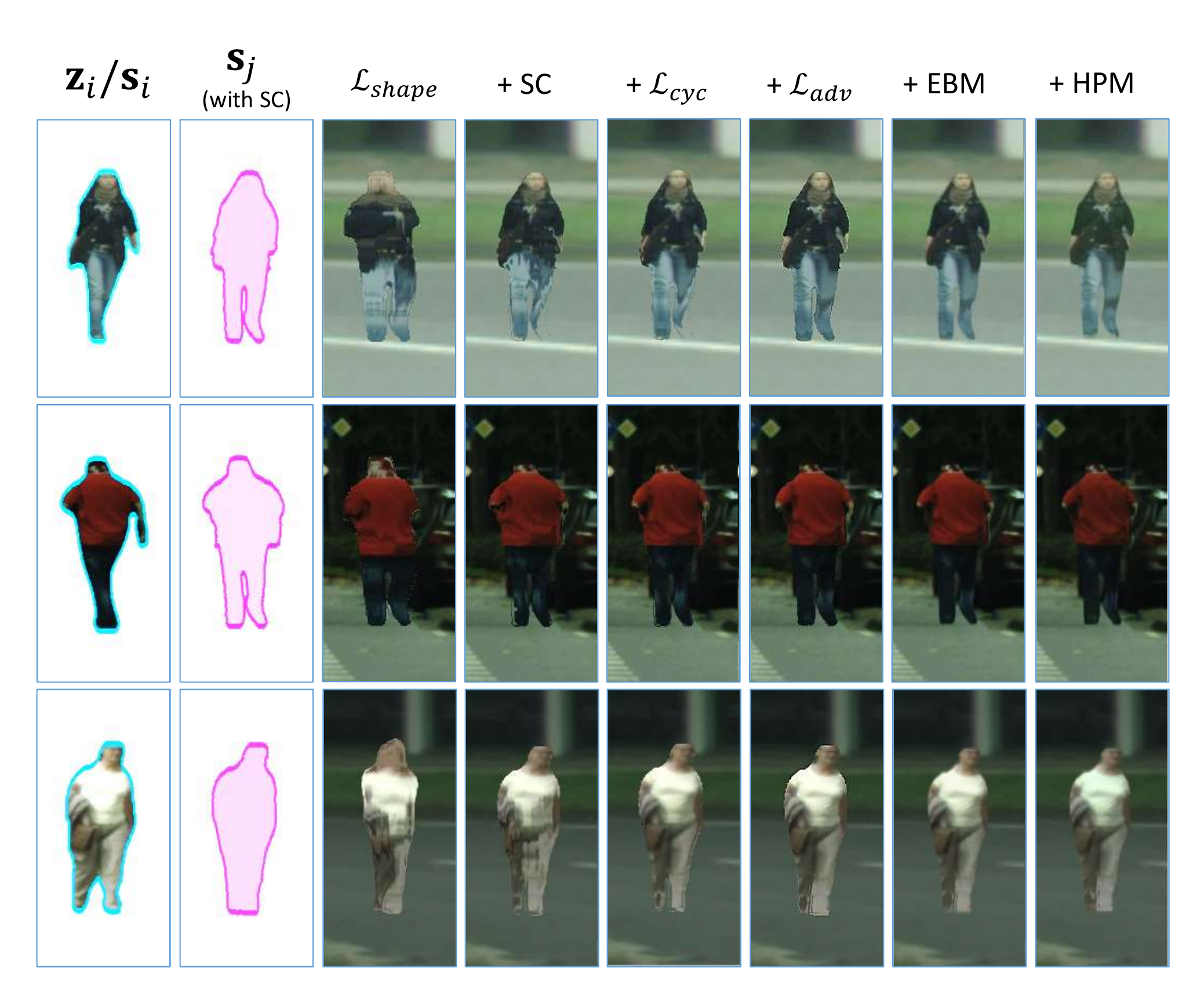}
\end{center}
   \caption{Effects of different components in the proposed STDA framework. ``SC'' means shape constraining operation; ``EBM'' means environment-aware blending map; ``HPM'' means hard positive mining. Best view in color.}
\label{fig:abl}
\end{figure*}

\begin{table*}[h]
\begin{center}
\resizebox{0.8\textwidth}{!}{
  \begin{tabular}{|M{1cm} | M{1cm} |M{1cm} | M{1cm} | M{1cm} | M{1cm} | M{1cm} || M{1cm} |}
  \hline
   baseline & $\mathcal{L}_{shape}$ & SC&  $\mathcal{L}_{cyc}$ &$\mathcal{L}_{adv}$ & EBM&HPM & MR\% \\
  \hline
  \ding{51} & & & & &&&10.77\\
  \hline
  \ding{51} & \ding{51}& & & & & &10.58\\
  \hline
  \ding{51} & \ding{51} &\ding{51} & & & &&10.31\\
  \hline
  \ding{51} & \ding{51}&\ding{51}& \ding{51}& & & &9.03\\
  \hline
  \ding{51} &\ding{51}& \ding{51}&\ding{51} &\ding{51}& & &8.56\\
  \hline
  \ding{51} &\ding{51}& \ding{51}&\ding{51} &\ding{51}&\ding{51} &&7.73\\
  \hline
  \ding{51} &\ding{51} & \ding{51}&\ding{51} &\ding{51}&\ding{51} &\ding{51}&7.49\\
  \hline
  \end{tabular}
}
\end{center}
\caption{Effects of different components in the proposed STDA framework on the selected validation set on Caltech dataset. ``SC'' means shape constraining operation; ``EBM'' means environment-aware blending map; ``HPM'' means hard positive mining.}
\label{tab:abl}
\end{table*}
\subsubsection{Qualitative Study}
We first evaluate the qualitative effects of different components in the STDA framework for the pedestrian generation task. In particular, we start the experiments from only using the shape-guided deformation supervised by $\mathcal{L}_{shape}$ for pedestrian generation. Then, we gradually add the shape-constraining operation, cyclic reconstruction loss $\mathcal{L}_{cyc}$, adversarial loss $\mathcal{L}_{adv}$, environment-aware blending map $\mathbf{e}(x,y)$, and hard positive mining loss $\mathcal{L}_{hpm}$ to help generate pedestrians. We present the effects of different components by generating pedestrians based on low-quality real pedestrian data in the Fig. \ref{fig:abl}. According to the presented results, we can observe that the quality of the generated pedestrians is progressively improved by introducing more components, demonstrating the effectiveness of the different components in STDA framework. More specifically, the shape constraining operation can first help the deformation operation produce less distorted pedestrians. Then, by adding the cyclic loss $\mathcal{L}_{cyc}$ and adversarial loss $\mathcal{L}_{adv}$, the obtained pedestrians become more realistic-looking in details. Subsequently, the introduced environment-aware blending map trained by $\mathcal{L}_{ebm}$ helps the transformed pedestrians better adapt into the background image patch. Lastly, the $\mathcal{L}_{hpm}$ can slightly change some appearance characteristics, such as illumination or color, to make the pedestrians less distinguishable from the environments, which actually further improve the pedestrian generation results.

\subsubsection{Quantitative Study}
To perform ablation studies, we split the training set of Caltech into one smaller training set and one validation set. More specifically, we collect the frames from the first four sets in the training as training images, while the frames from the last set are considered as validation images. We sample every 30-th frame in the overall dataset to set up the training/validation set. Note that this setting of training/validation set is ONLY used for ablation study.

Table \ref{tab:abl} presents the detailed results. We can find that each of the introduced component, including shape constraining operation (SC), cyclic loss ($\mathcal{L}_{cyc}$), adversarial loss ($\mathcal{L}_{adv}$), the environment-aware blending map (EBM), and the hard positive mining (HPM), can all contribute a promising average miss rate reduction. In particular, the cyclic and adversarial loss that helps better deform pedestrians and the environment-aware blending map that helps better adapt deformed pedestrians can both greatly boost the benefits of synthesized pedestrians on improving detection accuracy. The proposed hard positive mining scheme can further improve the detection accuracy, demonstrating its effectiveness in dataset augmentation. Based on the qualitative analysis as shown in Figure \ref{fig:abl}, we can further conclude that augmenting pedestrian datasets with more realistic-looking pedestrians can deliver better improvements on detection accuracy.

\section{Conclusions}
In this study, we present a novel shape transformation-based dataset augmentation framework to improve pedestrian detection. The proposed framework can effectively deform natural pedestrians into a different shape and can adequately adapt the deformed pedestrians into various background environments. Using low-quality pedestrian data available in the datasets, our proposed framework produces much more lifelike pedestrians than other cutting-edge data synthesis techniques. By applying the proposed framework on the two different well-known pedestrian benchmarks, \ie~ Caltech and CityPersons, we improve the baseline pedestrian detector with a great margin, achieving state-of-the-art performance on both of the evaluated benchmarks. 

%
%

\bibliographystyle{spbasic}      
\bibliography{egbib}   

\end{document}